\documentclass[sigconf]{acmart}
\AtBeginDocument{%
  \providecommand\BibTeX{{%
    Bib\TeX}}}
\setcopyright{none}
\renewcommand\footnotetextcopyrightpermission[1]{}
\authorsaddresses{}

\copyrightyear{2026}
\acmYear{2026}
\setcopyright{cc}
\setcctype{by}
\acmConference[SIGGRAPH Posters '26]{Special Interest Group on Computer Graphics and Interactive Techniques Conference Posters}{July 19--23, 2026}{Los Angeles, CA, USA}
\acmBooktitle{Special Interest Group on Computer Graphics and Interactive Techniques Conference Posters (SIGGRAPH Posters '26), July 19--23, 2026, Los Angeles, CA, USA}
\acmDOI{10.1145/3799825.3818750}
\acmISBN{979-8-4007-2548-7/2026/07}

\begin{teaserfigure}
  \includegraphics[width=\textwidth]{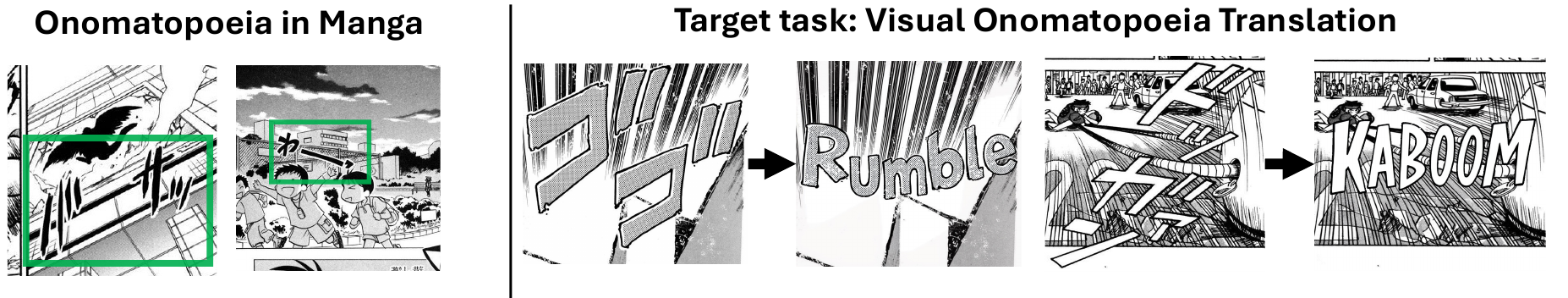}
  \caption{Left: Onomatopoeia appearing in Manga. From DualJustice, $\copyright$ Yusuke Takeyama, Arisa $\copyright$ Ken Yagami.
  Right: Our target task, which replaces Japanese onomatopoeia with the English one. From AppareKappore $\copyright$ Hirsohi Kanno, GarakutayaManta $\copyright$ Tatsuki Nouda}\label{fig:teaser}
\end{teaserfigure}
\begin{document}

\title{OnomatoBridge: Onomatopoeia Translation and Rendering Pipeline in Manga}


\author{Takara Taniguchi}
\author{Hideki Nakayama}
\email{hiroshi-tani@g.ecc.u-tokyo.ac.jp}
\affiliation{
  \institution{The University of Tokyo}
\country{Japan} 
}

\author{Wataru Shimoda}
\author{Kota Yamaguchi}
\affiliation{%
  \institution{CyberAgent}
\country{Japan} 
}

\begin{abstract}
Manga is a comic drawn by black and white paints gaining popularity around the world.
Onomatopoeia in Manga specifically appeals to the audience with its unique visual styles, which convey sound, motion, and emotion.
\textit{Visual onomatopoeia translation} requires the clean replacement of Japanese onomatopoeia with onomatopoeia in the other language while preserving their visual style.
Existing approaches often produce residual artifacts or style inconsistency when removing the Japanese onomatopoeia and rendering stylized English onomatopoeia. 
To approach these problems, we present OnomatoBridge, a filtering pipeline for visual onomatopoeia translation. 
We evaluate OnomatoBridge from Japanese to English on the Manga109 onomatopoeia dataset and compare it with baseline image editing models.
Experimental results show that the filtered outputs by the proposed method outperform those of conventional methods.
OnomatoBridge improves English text correctness by roughly 10 to 25 points and reduces residual Japanese text by about 20 to 50\% in relative terms.
\end{abstract}



\maketitle

\section{Introduction}

Manga is a comic with its black and white sequential artwork, gaining popularity around the world. 
A typical manga page includes character and scene illustrations, screen-tone effects, and dialogue placed within various panel arrangements. 
Prior research has investigated these unique characteristics of manga in tasks such as colorization~\cite{mangacolorization2021ouyang}, and inpainting~\cite{mangainpainting2021xie}.

\textit{Visual onomatopoeia translation}, which renders visualized onomatopoeia from Japanese to English while keeping visual styles shown in Fig.~\ref{fig:teaser}, is important for the localization of Manga since it is used to express sounds, emotions, and movement, deepening the reader's immersion. 
While recent research addresses the generation of Japanese onomatopoeia~\cite{taniguchi2025onomatogenonomatopoeiagenerationalphachannel} and the artistic typography~\cite{Vinker2023typography}, \textit{visual onomatopoeia translation} remains underexplored.
The inability of traditional methods to generate prompt-aligned onomatopoeia poses a challenge to the replacement process.
Conventional models~\cite{labs2025flux1kontextflowmatching, wu2025qwenimagetechnicalreport} can generate artifacts (i.e., residual Japanese onomatopoeia left after replacement) and English onomatopoeia unmatched with the original style.

\begin{figure*}[t]
  \centering
  \includegraphics[width=0.9\textwidth]{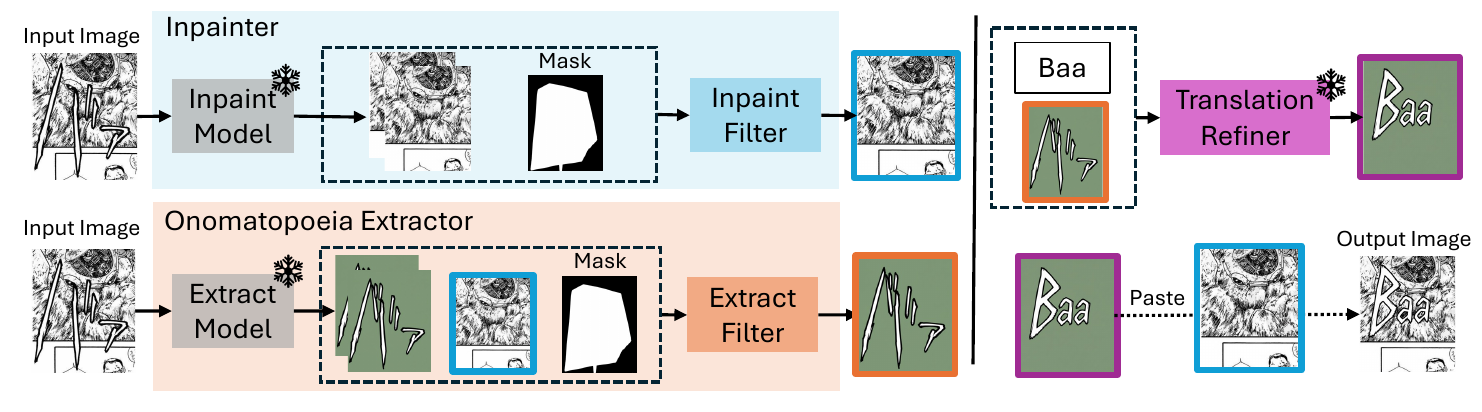}
\caption{Each component of OnomatoBridge. Right: Whereas the input image of OnomatoBridge is processed to inpaint the onomatopoeia, it is simultaneously processed to select the best extracted onomatopoeia. Left: The output of the onomatopoeia extractor is transformed to the English onomatopoeia and it is pasted to the output of inpainter. From EvaLady $\copyright$ Mio Iisen}\label{fig:method}
\end{figure*}

We construct OnomatoBridge, the filtering pipeline to replace Japanese onomatopoeia with English onomatopoeia to approach the difficulty of alignment failure against the prompt.
Our method replaces onomatopoeia through three stages: inpainting the original onomatopoeia, extracting the original onomatopoeia, and rendering the translated onomatopoeia, as illustrated in Fig.~\ref{fig:method}.
At each stage, the module generates multiple candidates and sequentially filters them to retain only those assessed as reliable, rather than falling back to low-quality outputs when no suitable candidate is found.
Experimental results show that the outputs after filtering as high quality by the proposed method outperform those of conventional methods.



\section{Methods}

Existing methods fail to minimize image artifacts, preserve stylistic consistency with the original Japanese, and maintain the readability of the generated English onomatopoeia.
OnomatoBridge tackles these problems with the inpainter, the onomatopoeia extractor, and the translation refiner.

\noindent\textbf{Task Definition.}
The input to OnomatoBridge is a manga image containing a Japanese onomatopoeia region together with its corresponding translation text and mask region $P$, and the output is a manga image in which the original Japanese onomatopoeia is removed and replaced with an English onomatopoeia.
Multiple images are generated and filtered during the process of OnomatoBridge.
We assume that the translated English text and the mask region are already given.

\noindent\textbf{Inpainter.}
Since the residual of the original Japanese onomatopoeia degrades the quality of visual onomatopoeia translation, the inpainter selects the candidate where the only original onomatopoeia is cleanly removed from multiple inpainted images.
An inpaint model, which is an image editing model prompted to inpaint the onomatopoeia, generates multiple images with different seed values.
Smaller pixel changes inside the mask indicate a failure to inpaint the onomatopoeia, whereas larger changes outside the mask also suggest that areas other than the onomatopoeia are mistakenly inpainted.
The inpaint filter scores multiple candidates based on the relative proportions of changed pixels inside and outside the mask, and selects the candidate with the highest score.
By comparing the original image $I$ and an inpainted image $I'$, we compute $S$ and $S'$, the proportions of pixels whose intensity difference exceeds a threshold inside and outside the mask $P$, respectively.
We reject candidates with insufficient in-mask change or excessive out-of-mask change by using the threshold, and choose the remaining candidate maximizing $S-S'$.

\noindent\textbf{Onomatopoeia Extractor.}
The onomatopoeia extractor selects the candidate that best preserves the original position and visual style of the source onomatopoeia from multiple images.
An extract model, which is also an image editing model, outputs the image where the onomatopoeia is extracted in the green background.
In the greenback image, the presence of non‑green pixels outside the polygon mask indicates that the onomatopoeia has not been correctly extracted at the original position; therefore, excluding such cases is important.
For each onomatopoeia-extracted image, we compute grayscale concentration ratios inside and outside the polygon region $P$, denoted by $s$ and $s'$, to determine whether the extracted onomatopoeia is confined to the original position.
Since the extracted onomatopoeia will later be used as a reference, it is important to evaluate whether its visual style has been correctly extracted.
We also calculate SSIM $M$ between the original image and a composite image formed by overlaying the extracted onomatopoeia on the selected inpainted image to verify that the extracted onomatopoeia is similar to the original visual styles.
We calculate the final score as $C = (s - s') + M$ to combine the two factors mentioned above.
Candidates whose final scores fall below a threshold are rejected, and the one with the highest $C$ is selected.

\noindent\textbf{Translation Refiner.}
The translation refiner is a module that converts Japanese onomatopoeia, extracted onto a green background by the extract filter, into readable English onomatopoeia.
Since a single-pass transformation can result in partially unreadable or unnatural English onomatopoeia, we perform image editing in two stages.
First, we run FLUX.1 Kontext Pro with the prompt \textit{"Replace all Japanese onomatopoeia with \{translated\_onomatopoeia\}. Keep the original font style and color."} to perform the initial replacement. 
Second, we run FLUX.1 Kontext Pro with the prompt \textit{"Edit this image to keep only the English text \{translated\_onomatopoeia\} with the same visual style."} as a refinement pass to clean residual artifacts.

\section{Experiment}
\begin{table*}[t]
\setlength{\tabcolsep}{16pt} 
\small
\caption{Quantitative comparison and user study results. NED (Normalized Edit Distance) measures text accuracy, while JPDet rate (Japanese Detection Rate) indicates the ratio of original Japanese text remaining after editing. Style, Residual, and Readability are subjective scores (1–5) evaluating artistic integration, background inpainting quality, and text legibility, respectively.}\label{tbl:quantitative}
\begin{tabular}{@{}lcc|ccc@{}}
\toprule
Model name           & NED $\uparrow$ & JPDet rate$\downarrow$ & Style $\uparrow$ & Residual $\uparrow$ & Readability $\uparrow$ \\ \midrule
Naïve baseline       & \textbf{0.8918}  & 0.5108             &  -     &           -        &       -      \\
Flux.1 Kontext~\cite{labs2025flux1kontextflowmatching}       & 0.5391  & 0.2772             & 2.151 & 3.548             & 3.014       \\
Qwen Image Edit~\cite{wu2025qwenimagetechnicalreport}      & 0.5673  & 0.4547             & 1.986 & 2.548             & 2.863       \\
Gemini Image 3 pro   & 0.7182  & 0.3949             & 2.945 & 1.944             & 3.945       \\
OnomatoBridge (ours) & 0.7826  & \textbf{0.2246}             & \textbf{3.098} & \textbf{3.946}             & \textbf{4.121}       \\ \bottomrule
\end{tabular}
\end{table*}
\begin{figure*}[htbp]
    \centering
    \includegraphics[width=0.9\textwidth]{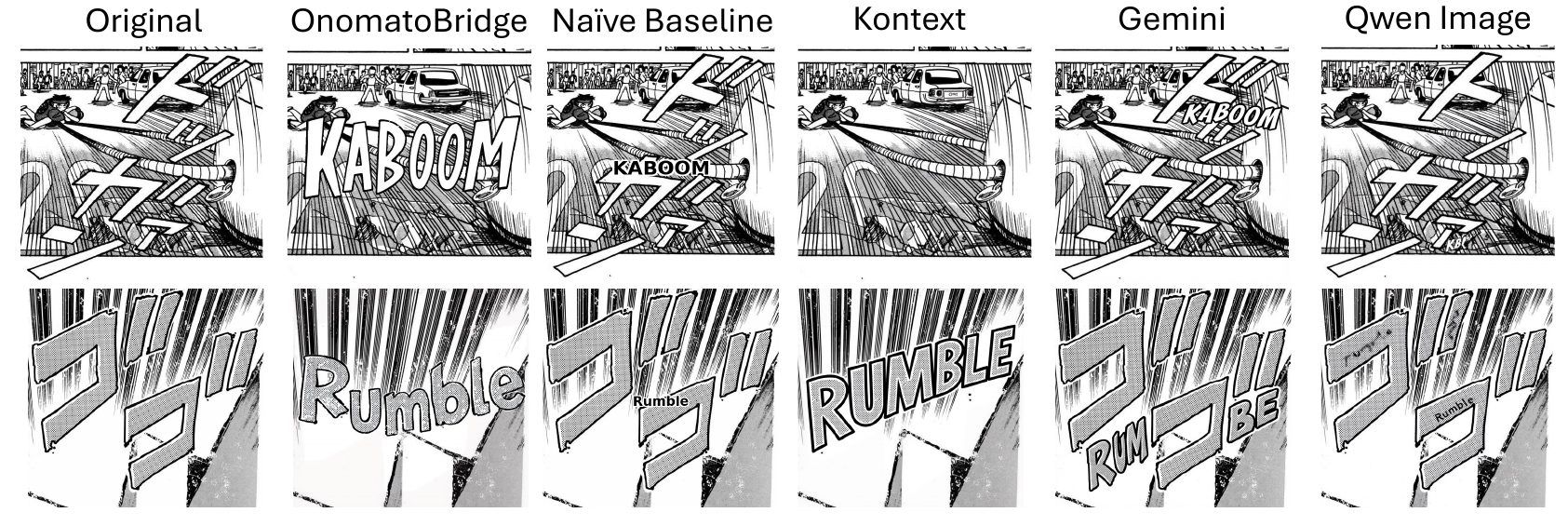}
    \caption{Quantitative comparison between OnomatoBridge and baselines. OnomatoBridge reflects the original style of Japanese onomatopoeia and cleanly removes the residuals.}
    \label{fig:qualitative}
\end{figure*}

We compare OnomatoBridge with the other image editing models by using quantitative metrics and a user study.

\noindent\textbf{Implementation detail.}
We use FLUX.1 Kontext~\cite{labs2025flux1kontextflowmatching} and Gemini-2.5-flash-image for an inpaint model and an extract model, respectively.
The inpaint filter takes four generated images and selects the best image, while we generate four candidates and choose the best image from them for the extract filter.

\noindent\textbf{Dataset.}
We use Manga 109 onomatopoeia~\cite{baek2022onomatopoeia} for the experiment. 
Manga 109 onomatopoeia includes the image of Japanese onomatopoeia with the corresponding Japanese text. 
To target large onomatopoeia that are visually eminent to readers, we use 384 Japanese onomatopoeia whose bounding boxes, enclosing the onomatopoeia polygons, are at least $300 \times 300$.
For the evaluation, we use 92 images after filtering by OnomatoBridge.
The text of English onomatopoeia is generated from GPT-4o with the input of the original Japanese onomatopoeia.
The mask $P$ covering the onomatopoeia is given by the dataset.

\noindent\textbf{Evaluation Metric.}
We evaluate VLM outputs using Normalized Edit Distance (NED) to assess the correctness of the generated English onomatopoeia, and additionally use VLM-based True/False judgments on the generated images to verify whether the original Japanese onomatopoeia has been cleanly removed.
We also conduct a user study with five people to ask whether the original style of Japanese onomatopoeia is kept, whether the residual artifact exists, and whether the generated onomatopoeia follows the original input of English text.
The user study is conducted for five people using a Likert scale from one to five.

\noindent\textbf{Baselines.}
We run FLUX.1 Kontext~\cite{labs2025flux1kontextflowmatching}, Qwen-Image Edit~\cite{wu2025qwenimagetechnicalreport}, and Gemini 3 Pro Image using the Japanese onomatopoeia image and a text prompt and a mask.
The naive baseline simply overlays the Arial English text onto the original image containing Japanese onomatopoeia.

\noindent\textbf{Results.}
Table~\ref{tbl:quantitative} indicates that our method outperforms competing models except for the naive baseline in both English onomatopoeia readability and removal of the original Japanese onomatopoeia.
We are evaluating the experimental results using these 92 English onomatopoeia, which is the number of onomatopoeia after filtering of OnomatoBridge.
The performance of NED shows that the naive baseline is quite readable but has low performance, since it just pastes the English Arial, which shows that these metrics can be hacked.
Shown in Fig.~\ref{fig:qualitative} and the user study of Table~\ref{tbl:quantitative}, our method can replace onomatopoeia, preserving the original style of Japanese onomatopoeia with fewer residual artifacts.

\section{Conclusion}
We present a pipeline that can replace onomatopoeia from Japanese to English, while preserving the original style of Japanese onomatopoeia.
A limitation of this study is the low filtering yield, and our method cannot reliably replace onomatopoeia that appears in background regions.
An end-to-end application is a good candidate for future study.

\bibliographystyle{ACM-Reference-Format}
\bibliography{main}

\end{document}